\documentclass[runningheads]{llncs}
\usepackage[utf8]{inputenc} 
\usepackage[T1]{fontenc}    
\usepackage{url}            
\usepackage{booktabs}       
\usepackage{amsfonts}       
\usepackage{xfrac}
\usepackage{microtype}      
\usepackage{xcolor}         

\usepackage{listings}
\usepackage{enumerate}
\usepackage{soul}
\usepackage{lipsum}
\usepackage{bbm}

\usepackage{graphicx, subfigure}
\usepackage{multirow, longtable, diagbox, setspace, rotating}
\usepackage{wrapfig, float}

\usepackage{amsmath, amssymb, amsfonts} 
\usepackage{mathtools}
\usepackage[ruled, vlined]{algorithm2e}
\usepackage{algpseudocode}

\usepackage{tikz}
\usetikzlibrary{arrows.meta, positioning, shapes.geometric, calc, decorations.pathmorphing}

\begin{document}
\title{RichMap: A Reachability Map Balancing \\ Precision, Efficiency, and Flexibility for \\ Rich Robot Manipulation Tasks}
\titlerunning{RichMap: Balancing Precision, Efficiency, and Flexibility for Reachability}
%

\author{Yupu Lu\inst{1}
\and
Yuxiang Ma\inst{2}
\and
Jia Pan\inst{1}\thanks{Corresponding author}
}
\authorrunning{Y. Lu et al.}
%
\institute{School of Computing and Data Science, The University of Hong Kong, HKSAR\\
\email{\{luyp16,panj\}@connect.hku.hk}\\
\and
Massachusetts Institute of Technology, Cambridge, USA\\
\email{yxma20@mit.edu}
}
\maketitle              
\begin{abstract}
This paper presents RichMap, a high-precision reachability map representation designed to balance efficiency and flexibility for versatile robot manipulation tasks. 
By refining the classic grid-based structure, we propose a streamlined approach that achieves performance close to compact map forms (e.g., RM4D) while maintaining structural flexibility. Our method utilizes theoretical capacity bounds on $\mathbb{S}^2$ (or $SO(3)$) to ensure rigorous coverage and employs an asynchronous pipeline for efficient construction. 
We validate the map against comprehensive metrics, pursuing high prediction accuracy ($>98\%$), low false positive rates (1$\sim$2\%), and fast large-batch query ($\sim$15 $\mu$s/query). 
We extend the framework applications to quantify robot workspace similarity via maximum mean discrepancy (MMD) metrics and demonstrate energy-based guidance for diffusion policy transfer, achieving up to 26\% improvement for cross-embodiment scenarios in the block pushing experiment.

\end{abstract}
\section{Introduction}

Reachability maps encode the set of reachable end-effector poses, serving as critical priors for robot manipulation tasks from base placement optimization~\cite{vahrenkamp2013robot} to motion planning~\cite{zhang2020novel}.
The classical spatial-rotational grid structure~\cite{zacharias2007capturing,zacharias2013capability} has been widely adopted, but can be inefficient concerning the curse of dimensionality. To address this issue, other existing implementations sacrifice the simplicity for compactness~\cite{han2021efficient,Rudorfer2025RM4D}.

In this work, we propose RichMap, a reachability map framework designed to optimize the trade-off between precision, efficiency, and flexibility:

\textbf{Effective theoretical analysis:} We adopt a minimal kinematic assumption, a freely rotating wrist joint, and derive capacity bounds for orientation storage via geodesic distance on $\mathbb{S}^2$ (or $SO(3)$). This analysis guides parameter selection (cell size, angular threshold) to balance precision and storage, enabling reachability prediction comparable to compact 4D maps with low false positive rates.

\textbf{Efficient construction framework:} We design an asynchronous accelerated pipeline that decouples CPU-based forward kinematics sampling from GPU-based batched insertion. The framework scales to massive datasets ($10^8$ poses) while maintaining high throughput as the map densifies and achieving averaged microsecond-level query latencies for large batches.

\textbf{Extended applications:} The explicit separation of spatial and rotational components enables novel applications beyond classic reachability queries. We demonstrate robot workspace similarity quantification via cell-wise maximum mean discrepancy (MMD) metrics, and energy-based guidance for diffusion policy transfer, bringing challenging cross-embodiment cases to near-source performance levels ($\sim$90\% vs.~95\% source baseline) in the block pushing task.

\section{Related Works}

Reachability analysis is for evaluating workspace capabilities in robotics. 
In the joint space, early approaches utilized manipulability~\cite{yoshikawa1985manipulability} to measure the determination of the manipulator posture in the workspace. This concept was later extended to account for joint limits and collisions~\cite{vahrenkamp2012manipulability,vahrenkamp2015representing}, providing local quality measures. Recent work also provides another way to analyze joint space dexterity given end-effector poses~\cite{yao2023enhanced}.

To provide global reachability information for the end-effector pose space, discrete map representations have been widely adopted. Capability maps~\cite{zacharias2007capturing} discretize the workspace into 3D grids to store reachability indices, such as the percentage of reachable orientations on a sphere~\cite{castelli2008fairly,porges2015reachability}. Higher-fidelity representations store detailed kinematic information within $SE(3)$ grids~\cite{cao2011accurate,zacharias2013capability}, enabling precise feasibility checks. It can also integrate manipulability metrics for richer functional analysis~\cite{xu2021optimal}. However, the curse of dimensionality often limits the resolution or completeness of these maps.

Recent research focuses on optimizing storage and query efficiency. Analytical approaches utilize mathematical structures of the kinematic manifold to reduce data redundancy~\cite{han2021efficient,quan2022dexterity,Rudorfer2025RM4D,cavelli2025modeling}. Alternatively, learning-based methods approximate reachability boundaries using neural networks~\cite{jiang2025graph} to achieve continuous representations. Nonetheless, implicit representations and complex map structure will create difficulties in fast and wide applications.
Our work addresses these limitations by proposing a balanced representation that maintains the simplicity and precision of explicit grid-based maps while leveraging theoretical guarantees and asynchronous generation pipeline to ensure high performance and flexibility.

\section{Methodology}
\label{sec:methodology}

By theoretically analyzing position discretization and orientation coverage, we design the reachability map to efficiently encode the robot's workspace capabilities, enabling fast queries while maintaining high accuracy.

\subsection{Problem Formulation}

Consider a robot manipulator with forward kinematics function $f_k: \mathbb{R}^n \rightarrow SE(3)$, which maps a joint configuration $\mathbf{j} \in \mathbb{R}^n$ to an end-effector pose $\mathbf{T} \in SE(3)$ represented as 7-dimensional vectors $\mathbf{p} = [\mathbf{t}, \mathbf{q}]^\top$, where $\mathbf{t} = (x, y, z)$ specifies the position and the unit quaternion $\mathbf{q} = (q_w, q_x, q_y, q_z)$ encodes the orientation with $q_w \geq 0$ for simplicity (due to the antipodal identification $\mathbf{q} \equiv -\mathbf{q}$).

The reachability map aims to efficiently answer the query: given a target pose $\mathbf{p}^*$, determine whether there exists a collision-free joint configuration $\mathbf{j}^*$ such that $f_k(\mathbf{j}^*) \approx \mathbf{p}^*$ within some tolerance.

\subsection{Grid-Based Workspace Discretization}

To enable efficient spatial queries, we discretize the robot's workspace $\mathcal{W} \subset \mathbb{R}^3$ into a uniform 3D grid with cell size $\Delta \in \mathbb{R}^+$. The workspace bounds are defined as $\mathcal{W} = [x_{\min}, x_{\max}] \times [y_{\min}, y_{\max}] \times [z_{\min}, z_{\max}]$. For any position $\mathbf{t}$, the corresponding grid cell index per dimension is computed as:
\begin{equation}
 i_d = \left\lfloor \frac{t_d - t_{d,\min}}{\Delta} \right\rfloor, d \in \{x, y, z\},
\end{equation}
where indices are clamped to satisfy $0 \leq i_d < n_d$, with $n_d = \lceil (t_{d,\max} - t_{d,\min}) / \Delta \rceil$ denoting the number of cells along dimension $d$. This discretization yields a total of $N_{\text{grid}} = n_x \cdot n_y \cdot n_z = O(n^3)$ grid cells, where $n$ characterizes the workspace size relative to the cell resolution. By optimizing inserted poses, we can achieve a efficient balance between high spatial precision of $\Delta = 0.02$ m. 

\subsection{Orientation Coverage with Geodesic Distance}

Within each grid cell, we store a compact set of reachable end-effector approach directions. Following the assumption in reachability analysis~\cite{Rudorfer2025RM4D} that \textit{the last wrist joint can rotate freely around 360 degrees with its axis aligned to the end-effector approach vector}, we represent orientations by the approach directions on the unit sphere $\mathbb{S}^2$ rather than full quaternions. 

This reduction from $SO(3)$ to $\mathbb{S}^2$ significantly reduces storage requirements while maintaining sufficient orientation coverage for most manipulation tasks. But noted in Section~\ref{subsec:workflow_acceleration}, our pipeline actually supports progressively relaxing this assumption through rotation range tracking during map construction.

For two unit direction vectors $\mathbf{v}_1, \mathbf{v}_2 \in \mathbb{S}^2$, the geodesic distance (great circle distance) on the sphere is:
\begin{equation}
d_{\mathbb{S}^2}(\mathbf{v}_1, \mathbf{v}_2) = \arccos(\mathbf{v}_1 \cdot \mathbf{v}_2),
\end{equation}
where the dot product gives the cosine of the angle between the two directions.

Using this metric, we apply the following insertion criterion: a new pose with approach direction $\mathbf{v}$ is added to grid cell $\mathcal{C}(\mathbf{t})$ if and only if:
\begin{equation}
\min_{\mathbf{v}' \in \mathcal{V}_{\mathcal{C}(\mathbf{t})}} d_{\mathbb{S}^2}(\mathbf{v}, \mathbf{v}') \geq \theta,
\end{equation}
where $\mathcal{V}_{\mathcal{C}(\mathbf{t})}$ denotes the set of directions already stored in the cell, and $\theta$ is the angular threshold. This criterion ensures that stored directions are well-distributed while avoiding redundancy.

To understand the storage requirements, we analyze the theoretical capacity of each cell. The insertion criterion requires that any two stored directions $\mathbf{v}_i, \mathbf{v}_j$ satisfy $d_{\mathbb{S}^2}(\mathbf{v}_i, \mathbf{v}_j) \geq \theta$. Consider an arbitrary direction $\mathbf{v}^*$ on $\mathbb{S}^2$: by the triangle inequality, its distances to any two stored directions must satisfy:
\begin{equation}
d_{\mathbb{S}^2}(\mathbf{v}^*, \mathbf{v}_i) + d_{\mathbb{S}^2}(\mathbf{v}^*, \mathbf{v}_j) \geq d_{\mathbb{S}^2}(\mathbf{v}_i, \mathbf{v}_j) \geq \theta.
\end{equation}
Consequently, no direction can lie within distance $\theta/2$ of two distinct stored directions simultaneously. This property justifies assigning each stored direction an exclusive spherical cap of radius $\theta/2$, ensuring non-overlapping packing regions while respecting the minimum separation constraint.

The surface area of $\mathbb{S}^2$ is $4\pi$, and an unit spherical cap of radius $\psi$ on $\mathbb{S}^2$ has area $2\pi(1 - \cos \psi)$. Therefore, the maximum number of directions that can be packed with minimum separation $\theta$ per cell satisfies:
\begin{equation}
N_{\text{max}} \cdot 2\pi\left(1 - \cos\frac{\theta}{2}\right) \leq 4\pi,
\end{equation}
yielding
\begin{equation}
\overline{N}_{\text{max}} \approx \frac{2}{1 - \cos(\theta/2)}.
\end{equation}
For $\theta = 0.1$ rad ($\approx 5.7^\circ$), this gives the upper bound $\overline{N}_{\text{max}} \approx 1600$ directions. 

\begin{figure}[t]
\centering
\begin{minipage}[c]{0.48\textwidth}
    \centering
    \begin{tikzpicture}[scale=0.7]
        \foreach \x in {0,1,2,3} {
            \foreach \y in {0,1,2,3} {
                \draw[gray!40] (\x,\y,0) -- (\x,\y,2);
            }
        }
        \foreach \x in {0,1,2,3} {
            \foreach \z in {0,1,2} {
                \draw[gray!40] (\x,0,\z) -- (\x,3,\z);
            }
        }
        \foreach \y in {0,1,2,3} {
            \foreach \z in {0,1,2} {
                \draw[gray!40] (0,\y,\z) -- (3,\y,\z);
            }
        }
        
        \draw[dashed, blue, very thick] (1,1,0) -- (2,1,0);
        \draw[blue, very thick] (2,1,0) -- (2,2,0);
        \draw[blue, very thick] (2,2,0) -- (1,2,0);
        \draw[dashed, blue, very thick] (1,2,0) -- (1,1,0);
        \draw[blue, very thick] (1,1,1) -- (2,1,1) -- (2,2,1) -- (1,2,1) -- cycle;

        \draw[dashed, blue, very thick] (1,1,0) -- (1,1,1);
        \draw[blue, very thick] (2,1,0) -- (2,1,1);
        \draw[blue, very thick] (2,2,0) -- (2,2,1);
        \draw[blue, very thick] (1,2,0) -- (1,2,1);
        
        \node at (3.6,3.6,3.6) {\small $\mathcal{C}_{i,j,k}$};
        
        \draw[->, red, thick] (1.5,1.5,0.5) -- ++(0.3,0.3,0.2);  
        \draw[->, red, thick] (1.5,1.5,0.5) -- ++(-0.4,0.15,0.2); 
        \draw[->, red, thick] (1.5,1.5,0.5) -- ++(0.1,-0.3,0.3); 
        \node[red] at (2.3,1.3,0.3) {\tiny $\{\mathbf{v}_k\}$};

        \draw[->, thick] (-0.5,0,0) -- (3.5,0,0) node[right] {$y$};
        \draw[->, thick] (0,-0.5,0) -- (0,3.5,0) node[above] {$z$};
        \draw[->, thick] (0,0,-0.5) -- (0,0,2.5) node[left] {$x$};
        
        \draw[<->, thick] (0.25,-0.8,0) -- (1.25,-0.8,0);
        \node at (0.75,-1.1,0) {\small $\Delta$};
    \end{tikzpicture}
\end{minipage}
\hfill
\begin{minipage}[c]{0.48\textwidth}
    \centering
    \begin{tikzpicture}[scale=0.8]
        \pgfmathsetmacro{\xa}{2.5*cos(55)*cos(30)}
        \pgfmathsetmacro{\ya}{2.5*sin(55)*cos(30)}
        \pgfmathsetmacro{\xb}{2.5*cos(80)*cos(35)}
        \pgfmathsetmacro{\yb}{2.5*sin(80)*cos(35)}
        
        \begin{scope}
            \clip (-2.5,0) arc (180:0:2.5cm) -- (2.5,0) -- cycle;
            \shade[ball color=blue!10!white, opacity=0.3] (0,0) circle (2.5cm);
        \end{scope}
        
        \draw[thick] (-2.5,0) -- (2.5,0);
        
        \draw[thick] (-2.5,0) arc (180:0:2.5cm);
        
        \begin{scope}
            \clip (-2.5,0) rectangle (2.5,3);
            
            \draw[red!20, fill=red!10, opacity=0.5] (\xa,\ya) circle (0.9cm);
            \draw[red!20, fill=red!10, opacity=0.5] (\xb,\yb) circle (0.9cm);
            
            \draw[blue!40, dashed, thick] (\xa,\ya) circle (0.45cm);
            \draw[blue!40, dashed, thick] (\xb,\yb) circle (0.45cm);
            
            \draw[->, red, ultra thick] (0,0) -- (\xa,\ya);
            \filldraw[red] (\xa,\ya) circle (3pt);
            
            \draw[->, red, ultra thick] (0,0) -- (\xb,\yb);
            \filldraw[red] (\xb,\yb) circle (3pt);
            
            \foreach \angle/\elev in {20/25, 110/35, 150/40} {
                \pgfmathsetmacro{\x}{2.5*cos(\angle)*cos(\elev)}
                \pgfmathsetmacro{\y}{2.5*sin(\angle)*cos(\elev)}
                \draw[->, gray, thick] (0,0) -- (\x,\y);
                \filldraw[gray] (\x,\y) circle (2pt);
            }
        \end{scope}
        
        \node[red] at (\xa+0.3,\ya-0.7) {$\mathbf{v}_1$};
        \node[red] at (\xb-0.5,\yb-0.7) {$\mathbf{v}_2$};
        
        \draw[<->, thick, green!50!black] (0.4,0.5) arc (55:80:0.8cm);
        \node[green!50!black] at (0.4,0.9) {$\theta$};
        
        \node[blue!70!black] at (\xa+0.8,\ya-0.4) {\small $\theta/2$};
        \node[red!70!black] at (\xa+0.8,\ya+0.2) {\small $\theta$};
    \end{tikzpicture}
\end{minipage}
\vspace{-0.1cm}
\caption{Reachability map representation. Left: 3D workspace discretization with cell size $\Delta$. Each cell stores approach direction vectors $\{\mathbf{v}_k\}$ (red arrows). Right: Orientation coverage via spherical cap packing. Directions $\mathbf{v}_1, \mathbf{v}_2$ maintain separation $\theta$ (green arc), with exclusive $\theta/2$-radius caps (blue dashed) ensuring non-overlapping packing.}
\label{fig:grid_and_packing}
\vspace{-0.2cm}
\end{figure}
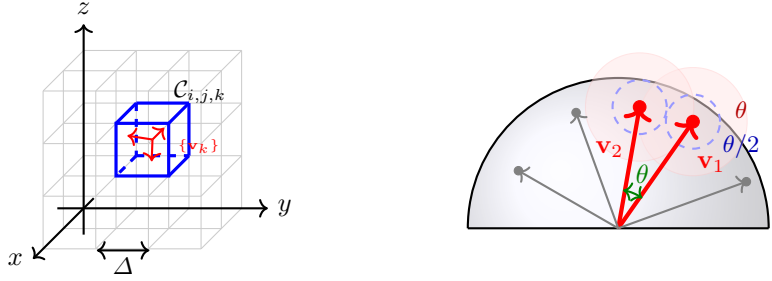

In practical construction with robots such as UR5e or Franka Panda, the average number of stored directions per cell is around $K =$ 300--500 and cell coverage rate 35--50$\%$. For cell size $\Delta = 0.02$ m, this yields approximately $2\times 10^8$ total stored poses, comparable to previous works and pursuing finer spatial resolution ($\Delta = 0.05$ m in RM4D~\cite{Rudorfer2025RM4D} and Capability Map~\cite{zacharias2013capability}).

\subsubsection{Full quaternion representation.} Naturally, the framework can store full $SO(3)$ quaternions. The geodesic distance corresponds to the rotation angle magnitude $d_{SO(3)}(\mathbf{q}_1, \mathbf{q}_2) = 2\arccos(|\mathbf{q}_1 \cdot \mathbf{q}_2|)$. The total volume of $SO(3)$ with the canonical metric is $\pi^2$. The volume of an unit ball with rotational radius $\psi$ in $SO(3)$ is given by $V(\psi) = \pi(\psi - \sin \psi)$. Using sphere packing arguments with an exclusive cap radius of $\theta/2$, we obtain:
\begin{equation}
\overline{N}_{\text{max}} \approx \frac{\pi^2}{\pi(\theta/2 - \sin(\theta/2))} = \frac{\pi}{\theta/2 - \sin(\theta/2)}.
\end{equation}
With $\theta = 0.4$ rad ($\approx 22.9^\circ$), this yields $\overline{N}_{\text{max}} \approx 2360$ quaternions per cell. This captures complete orientation with a clear theoretical bound, which echoes the map setting ($30^\circ$) used in the Capability Map \cite{zacharias2013capability}. 

\subsection{Map Construction and Workflow Acceleration}
\label{subsec:workflow_acceleration}

We construct the reachability map through forward kinematics sampling, inserting poses only when they satisfy the geodesic distance criterion. The query operation locates the target cell, finds the nearest stored direction, and returns the associated joint configuration if within threshold $\theta$, enabling dual purposes: fast reachability checking and approximate IK solving with $O(K)$ complexity per cell. 

\begin{algorithm}[h]
\caption{Asynchronous Reachability Map Construction}
\label{alg:construction}
\SetAlgoLined
\KwIn{Robot model, workspace $\mathcal{W}$, $\Delta$, $\theta$, target $N_{\text{target}}$}
\KwOut{Grid-based reachability map $\mathcal{G}$}
Initialize empty grid $\mathcal{G}$; $N_{\text{inserted}} \leftarrow 0$\;
\While{$N_{\text{inserted}} < N_{\text{target}}$ and Inserted Rate > Threshold}{
    $\{\mathbf{j}_i, \mathbf{p}_i\}_{i=1}^{B} \leftarrow$ Sample\_FK\_Batch()\;
    \For{each cell $c$ with new poses $\{\mathbf{v}_i\}$}{
        Check $d_{\mathbb{S}^2}(\mathbf{v}_i, \mathbf{v}')$ vs old data (Phase 1)\;
        Resolve conflicts within new poses (Phase 2)\;
        Insert valid poses; $N_{\text{inserted}} \leftarrow N_{\text{inserted}} + |\text{valid}|$\;
    }
}
\Return $\mathcal{G}$\;
\end{algorithm}
\vspace{-0.4cm}

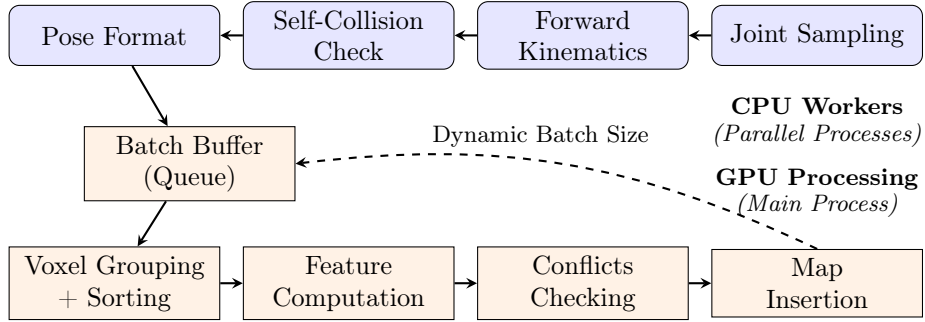
\begin{figure}[h]
\centering
\begin{tikzpicture}[
    node distance=0.6cm and 0.3cm,
    box/.style={rectangle, draw, rounded corners, minimum width=2.8cm, minimum height=0.8cm, align=center, fill=blue!10},
    process/.style={rectangle, draw, minimum width=2.8cm, minimum height=0.8cm, align=center, fill=orange!10},
    arrow/.style={->, >=stealth, thick}
]
    \node[box] (sample) {Joint Sampling};
    \node[box, left=of sample] (fk) {Forward\\Kinematics};
    \node[box, left=of fk] (collision) {Self-Collision\\Check};
    \node[box, left=of collision] (format) {Pose Format};
    
    \node[process, below=0.8cm of format, xshift=1cm] (buffer) {Batch Buffer\\(Queue)};
    
    \node[process, below=of buffer, xshift=-1cm] (group) {Voxel Grouping\\+ Sorting};
    \node[process, right=of group] (feature) {Feature\\Computation};
    \node[process, right=of feature] (check) {Conflicts\\Checking};
    \node[process, right=of check] (insert) {Map\\Insertion};
    
    \draw[arrow] (sample) -- (fk);
    \draw[arrow] (fk) -- (collision);
    \draw[arrow] (collision) -- (format);
    \draw[arrow] (format) -- (buffer);
    
    \draw[arrow] (buffer) -- (group);
    \draw[arrow] (group) -- (feature);
    \draw[arrow] (feature) -- (check);
    \draw[arrow] (check) -- (insert);
    
    \node[below=0.3cm of sample, align=center] {\small\textbf{CPU Workers}\\[-2pt]\small\textit{(Parallel Processes)}};
    \node[above=0.3cm of insert, align=center] {\small\textbf{GPU Processing}\\[-2pt]\small\textit{(Main Process)}};
    
    \draw[arrow, dashed, bend right=16] (insert.north) to node[above, xshift=-0.3cm, yshift=0.1cm] {\small Dynamic Batch Size} (buffer.east);
    
\end{tikzpicture}
\caption{Asynchronous pipeline architecture. Multiple CPU worker processes continuously generate collision-free poses in parallel while the main process executes GPU-accelerated batch insertions. The inter-process queue decouples the two stages, and batch size adjusts dynamically based on insertion throughput.}
\label{fig:pipeline}
\vspace{-0.2cm}
\end{figure}

\subsubsection{Asynchronous Pipeline Architecture.} Targeting massive datasets ($10^7$--$10^8$ poses), sequential map building can take 10+ days. To address this, we design an asynchronous producer-consumer pipeline where multiple CPU worker processes generate collision-free poses in parallel while the main process executes GPU-accelerated batch insertions. Figure~\ref{fig:pipeline} illustrates this workflow:

\textbf{CPU Generation Stage:} Multiple worker processes (typically 4-16) operate independently and in parallel, each executing four sequential operations: (1) random joint sampling within robot-specific limits, (2) forward kinematics evaluation, (3) self-collision detection via physics simulation, and (4) pose formatting to $\mathbf{p}$ in SE(3) space. The workers share no state and communicate only through a thread-safe queue, enabling linear scaling with CPU core count while maintaining a steady stream of collision-free poses.

\textbf{GPU Insertion Stage:} The main process vectorizes batches ($10^6$--$10^8$ poses). As the map densifies and insertion rate drops, batch size automatically increases to maintain GPU utilization. This adaptive batching strategy ensures consistent throughput of the construction process.

\subsubsection{Batched GPU Insertion.}
Sequential insertion becomes prohibitive as map density increases. Our batched strategy processes large chunks concurrently:
\begin{enumerate}
    \item \textbf{Voxel Grouping:} Sort incoming poses by destination cell index (stable sort preserves temporal order), creating groups for batch processing.
    \item \textbf{Feature Caching:} Compute and cache 3D axis vectors from quaternions in parallel, avoiding redundant computation during conflicts checks.
    \item \textbf{Two-Phase Conflicts Detection:}
    \begin{itemize}
        \item \textit{Phase 1 (New vs Old):} Batch matrix multiplication (BMM) computes geodesic distances between new candidates and existing cell data.
        \item \textit{Phase 2 (New vs New):} Greedy selection with geodesic distances checking and stable sorting resolves conflicts within new candidates.
    \end{itemize}
\end{enumerate}


\subsubsection{Rotation Range Tracking.}
The joint configurations sampled during generation enable progressive relaxation of the free wrist rotation assumption. For each inserted pose with approach direction $\mathbf{v}_i$, we maintain a set of observed wrist rotation angles $\Phi_i = \{[\phi_{1, \min}^i, \phi_{1, \max}^i], [\phi_{2, \min}^i, \phi_{2, \max}^i], \ldots\}$ based on joint configurations that share similar approach directions.

When a generated pose is rejected due to falling within the $\theta$-neighborhood of existing pose $i$, we can augment the rotation set:
\begin{equation}
\Phi_i \leftarrow \Phi_i \cup \{[\phi_{\text{new}, \min}, \phi_{\text{new}, \max}]\},
\end{equation}
where $[\phi_{\text{new}, \min}, \phi_{\text{new}, \max}]$ is the wrist angle range of the rejected pose. 

Section~\ref{subsec:exp1} shows the final insertion rate stabilizes around 6--7\% due to orientation redundancy. This 15$\times$ rejection ratio provides substantial opportunities to refine rotation coverage: for each stored direction, we observe roughly 15 distinct wrist angles from rejected poses, progressively approaching complete $SO(3)$ representation.

\subsection{Design Advantages}
\label{subsec:advantages}

Our framework optimizes for high precision, efficiency, and flexibility properties, balancing the lightweight nature of RM4D~\cite{Rudorfer2025RM4D} with the comprehensive utility of the Capability Map~\cite{zacharias2013capability}.

\subsubsection{High precision, minimal assumptions, and theoretical capacity bounds.}
We adopt a single kinematic assumption—a freely rotating wrist joint—from RM4D to simplify the general $\mathbb{R}^3+\mathbb{S}^2$ map structure. The following rigorous mathematical analysis of orientation capacity establishes a theoretical limit on the number of directions per cell, guiding settings selection for higher precision. 

Furthermore, rotation range tracking (section \ref{subsec:workflow_acceleration}) provides a way to progressively release this assumption, approaching $SO(3)$ coverage rather than directly discretize it.

\subsubsection{Flexibility through spatial-rotational separation.} 
We consciously elect to maintain a separated data structure (spatial grid and orientation set) typical of Capability Maps, rather than the highly compact representation of RM4D. 
Though incurring greater memory and computational costs, it affords the flexibility to empower diverse applications and extensions (Section \ref{subsec:applications}). 

Given the rapid expansion of modern hardware resources, we consider this cost a worthwhile trade-off for broader applicability and extensibility.

\subsubsection{Efficiency via asynchronous pipeline.} Our framework leverages an asynchronous producer-consumer pipeline to maximize throughput:      
\begin{enumerate}[(1)]
    \item This pipeline supports the decoupling of CPU-based FK sampling and GPU-based batched insertion. This parallelism ensures that map construction is limited only by the aggregate throughput, not sequential latency.
    \item FK evaluation is significantly faster than IK solving, and joint limits and collision constraints can be enforced during sampling without adjustments. Insertion is also accelerated through vectorized GPU processing.
    \item Joint configurations from sampling enable rotation range tracking. We can utilize the surplus of generated-but-rejected poses to refine orientation limits around stored approach directions.
\end{enumerate}

\subsection{Applications}
\label{subsec:applications}

The modular separation of spatial and rotational components enables several practical extensions.

\subsubsection{Map Compression.} For memory-constrained scenarios, an optional compression scheme is 
to represent each cell's orientation coverage using a single scalar value. Specifically, for each grid cell $c$, we generate $M$ quasi-uniform samples $\{\mathbf{v}_j^s\}_{j=1}^M$ on $\mathbb{S}^2$ using Fibonacci lattice sampling and compute the coverage fraction as:
\begin{equation}
\rho_c = \frac{1}{M} \sum_{j=1}^M \mathbbm{1}\left[\min_{\mathbf{v} \in \mathcal{V}_c} d_{\mathbb{S}^2}(\mathbf{v}_j^s, \mathbf{v}) < \theta\right],
\end{equation}
where $\mathbbm{1}[\cdot]$ denotes the indicator function. This compression reduces storage requirements from $O(N_{\text{grid}} \cdot K \cdot 3)$ floats for explicit direction storage to merely $O(N_{\text{grid}})$ floats for coverage fractions, which connects to implementations in~\cite{zacharias2007capturing,cao2011accurate,porges2015reachability}.


\subsubsection{Classic grasp-related tasks.} As a general reachability representation of the capability map~\cite{zacharias2013capability}, our framework can naturally support classic tasks such as reachability-aware mobile base placement~\cite{vahrenkamp2013robot,makhal2018reuleaux,kluge2021mobile}, feasible poses filtering for grasping~\cite{burget2015stance,jiang2022learning}, and motion planning with reachability priors~\cite{zhang2020novel,jauhri2022robot,sandakalum2022motion}.

\subsubsection{Robot similarity analysis.} The grid-based representation facilitates quantitative comparison of workspace capabilities across different robot models. 
We employ a cell-wise similarity metric based on Maximum Mean Discrepancy (MMD)~\cite{gretton2006kernel,gretton2012kernel} with a multi-scale Radial Basis Function (RBF) kernel. Given two robots with reachability grids $\mathcal{G}_A$ and $\mathcal{G}_B$ sharing identical spatial discretization, we compute for each corresponding cell pair $c$:
\begin{equation}
\text{MMD}^2(\mathcal{P}_A^c, \mathcal{P}_B^c) = \mathbb{E}_{AA} - 2\mathbb{E}_{AB} + \mathbb{E}_{BB},
\end{equation}
where $\mathbb{E}_{AA} = \mathbb{E}[k(\mathbf{p}_A, \mathbf{p}'_A)]$, $\mathbb{E}_{AB} = \mathbb{E}[k(\mathbf{p}_A, \mathbf{p}_B)]$, and $\mathbb{E}_{BB} = \mathbb{E}[k(\mathbf{p}_B, \mathbf{p}'_B)]$ denote the expected kernel similarities. The RBF kernel employs multi-scale bandwidth with $\gamma_m = 2^{m - M/2}$ for $m = 1, \ldots, M$ to capture similarities at different scales.

The resulting similarity grid $\mathcal{S}_{A \leftrightarrow B}$ quantifies workspace capability correspondence, with lower MMD values indicating greater functional similarity.

\subsubsection{Learning-based Manipulation Tasks.} We also integrate the robot similarity metric above into diffusion policy~\cite{chi2023diffusionpolicy,janner2022planning}. This metric supports energy-based guidance, enabling applications in policy transfer, as demonstrated in Section~\ref{subsec:exp3}.

\section{Experiments}
\label{sec:experiments}

As discussed in Section~\ref{subsec:advantages}, our reachability map is designed to achieve prediction performance comparable to the compact RM4D~\cite{Rudorfer2025RM4D} while retaining the structural flexibility of the Capability Map~\cite{zacharias2013capability}. To validate this design, we first benchmark our method against these two counterparts. We then demonstrate its practical utility through applications in robot similarity analysis and energy-based guidance for diffusion policy transfer.

\subsection{Reachability Map Verification}
\label{subsec:exp1}

\subsubsection{Experimental Setup.}
We conduct experiments on four widely-used robot manipulators: Franka Panda, KUKA iiwa7, Universal Robots UR5e, and Kinova Gen3. 

For verification, we generate 2 million test poses per robot with 50\% reachable samples and 50\% unreachable candidates. The latter comprises 10\% poses outside the workspace radius, 10\% below the ground plane, and 30\% with random perturbations in free space. All candidates are verified with 100 solution attempts per pose using IKFlow, a flow-based IK solver that can rapidly sample and optimize batch IK solutions for a given end-effector pose~\cite{ames2022ikflow}. A pose is labeled as feasible if at least one valid IK solution with no self-collision exists.

We construct reachability maps with workspace bounds of $[-1, 1] \times [-1, 1] \times [-1+\triangle_\text{z}, 1+\triangle_\text{z}]$ meters (0.2 or 0.3 adjusted for robot base offset) and geodesic distance threshold $\theta = 0.1$ rad. 
To investigate the effect of spatial resolution on prediction accuracy, we test two cell sizes: $\Delta = 0.05$ m (comparable to RM4D~\cite{Rudorfer2025RM4D}) and $\Delta = 0.02$ m (corresponding to $15.6\times$ more grid cells, $40^3 \rightarrow 100^3$). Map building stops when batch insertion rate drops below $1\%$. Every map can be built within 1 day with a superior consumer-level GPU.

\subsubsection{Evaluation Metrics.}
We monitor map construction every 1 million poses inserted, evaluating four prediction metrics:
\begin{itemize}
    \item \textbf{Accuracy:} Overall correct prediction rate
    \item \textbf{TPR (True Positive Rate):} Proportion of feasible poses correctly predicted as reachable
    \item \textbf{FPR (False Positive Rate):} Proportion of infeasible poses incorrectly predicted as reachable  
    \item \textbf{IR (Insertion Rate):} Proportion of generated poses successfully inserted into the map 
    \item \textbf{Query Time:} Average time ($\mu$s) per reachability query, averaged over a batch of test poses (2e6 if not specified).
\end{itemize}

\subsubsection{Results.}
We first compare our method against two counterparts at $\Delta=0.05$ m: RM4D~\cite{Rudorfer2025RM4D}, which uses compact 4D storage, and Capability Map (CM)~\cite{zacharias2013capability}, which stores full $SO(3)$ orientations (optimized with our framework, $\theta=0.4$ rad to match insertion level). Table~\ref{tab:counterpart_comparison} summarizes the comparison.

\begin{table}[h]
\centering
\caption{Comparison of Map Performance and Query Efficiency against Counterparts ($\Delta=0.05$ m, $\theta=0.1/0.4$ rad).}
\label{tab:counterpart_comparison}
\setlength{\tabcolsep}{2.5pt}
\small
\begin{tabular}{lcccccccccccc}
\toprule
\multirow{2}{*}{Method} & \multicolumn{3}{c}{Panda} & \multicolumn{3}{c}{IIWA7} & \multicolumn{3}{c}{UR5e} & \multicolumn{3}{c}{Kinova3} \\
\cmidrule(lr){2-4} \cmidrule(lr){5-7} \cmidrule(lr){8-10} \cmidrule(lr){11-13}
& Ins. & Acc. & FPR & Ins. & Acc. & FPR & Ins. & Acc. & FPR & Ins. & Acc. & FPR \\
\midrule
RM4D & 1.1M & .980 & .035 & 1.1M & .982 & .035 & 1.5M & .984 & .058 & 1.0M & .982 & .045 \\
CM & 10.7M & .974 & .052 & 9.8M & .972 & .058 & 16.4M & .973 & .082 & 9.1M & .972 & .063 \\
Ours & 11.3M & .975 & .037 & 10.2M & .978 & .036 & 18.2M & .978 & .063 & 10.0M & .978 & .046 \\
\midrule
\multicolumn{13}{c}{\textbf{Query Efficiency ($\mu$s/query) with Varying Batch Sizes}} \\
\midrule
\multirow{2}{*}{Method} & \multicolumn{3}{c}{Panda} & \multicolumn{3}{c}{IIWA7} & \multicolumn{3}{c}{UR5e} & \multicolumn{3}{c}{Kinova3} \\
\cmidrule(lr){2-4}\cmidrule(lr){5-7}\cmidrule(lr){8-10}\cmidrule(lr){11-13}
& 1e2 & 1e4 & 1e6 & 1e2 & 1e4 & 1e6 & 1e2 & 1e4 & 1e6 & 1e2 & 1e4 & 1e6 \\
\midrule
RM4D & 41.9 & 49.0 & 45.6 & 56.2 & 49.7 & 45.2 & 40.7 & 43.5 & 46.4 & 47.8 & 44.9 & 46.0 \\
CM & 78.6 & 38.3 & 2.7 & 82.4 & 43.0 & 2.7 & 71.2 & 46.1 & 3.0 & 78.6 & 39.0 & 2.4 \\
Ours & 91.4 & 45.3 & 2.5 & 82.0 & 47.7 & 2.9 & 94.1 & 50.2 & 3.0 & 76.1 & 43.8 & 2.7 \\
\bottomrule
\end{tabular}
\vspace{0.3cm}
\end{table}


\begin{table}[h]
\centering
\caption{Reachability Map Construction Results ($\theta=0.1$ rad).}
\label{tab:ikflow_verification}
\setlength{\tabcolsep}{4pt}
\small
\begin{tabular}{lccccccccc}
\toprule
Robot & $\Delta$ (m) & Inserted & Occ. & 
$\sfrac{\text{Avg}}{\text{Cell}}$ 
& Acc. & TPR & \textbf{FPR} &
$\text{IR}_\text{whole}$ & Time \\
\midrule
Panda & 0.05 & 11.3M & 35.2\% & 503 & 0.975 & 0.982 & 0.037 & 6.4\% & 1.2 \\
IIWA7 & 0.05 & 10.2M & 43.5\% & 367 & 0.978 & 0.987 & 0.036 & 6.0\% & 1.3 \\
UR5e & 0.05 & 18.2M & 52.7\% & 541 & 0.978 & 0.992 & 0.063 & 5.5\% & 1.3 \\
Kinova3 & 0.05 & 10.0M & 34.5\% & 454 & 0.978 & 0.991 & 0.046 & 5.4\% & 1.2 \\
\midrule
Panda & 0.02 & 164.1M & 32.8\% & 501 & 0.982 & 0.981 & 0.016 & 6.8\% & 13.5 \\
IIWA7 & 0.02 & 145.1M & 40.5\% & 358 & 0.984 & 0.984 & 0.015 & 6.7\% & 15.2 \\
UR5e & 0.02 & 260.0M & 49.5\% & 526 & 0.984 & 0.989 & 0.028 & 7.1\% & 15.1 \\
Kinova3 & 0.02 & 141.1M & 32.2\% & 439 & 0.984 & 0.986 & 0.019 & 7.1\% & 13.2 \\
\bottomrule
\end{tabular}
\end{table}

The bottom section of Table~\ref{tab:counterpart_comparison} details the query speed across varying batch sizes. Simplified by theoretical assumptions, RM4D achieves comparable accuracy with $\sim$10$\times$ fewer poses and exhibits relatively constant query time due to CPU-based sequential processing. As a comparison, our GPU-accelerated implementation (Ours and CM) shows significant speedup for large batches ($\sim$2-3 $\mu$s/query for $N=10^6$) despite 10$\times$ larger map size, highlighting the efficiency of our pipeline for bulk queries typical in sampling-based planning or learning. On the other hand, our method achieves \textbf{FPR} within 0.5\% of RM4D (0.037 vs. 0.035 for Panda) while consistently outperforming CM. With prediction accuracy close to the compact map form (RM4D) and accelerated batched query speed, \textbf{our method offers a favorable balance between efficiency and precision}. 

\begin{figure}[htbp]
\centering
\includegraphics[width=\textwidth]{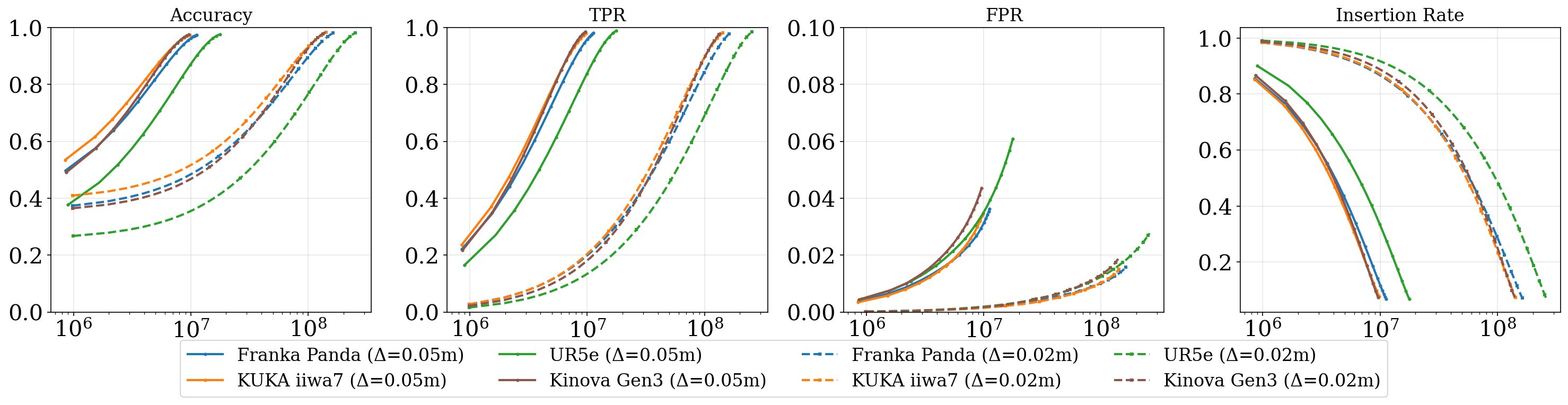}
\caption{Prediction metrics evolution during reachability map construction for four robots at two spatial resolutions ($\Delta=0.05$ m solid lines, $\Delta=0.02$ m dashed lines). All robots show rapid initial improvement followed by gradual saturation. Accuracy converges to $>97\%$ for both resolutions, demonstrating reliable feasibility prediction. The finer resolution ($\Delta=0.02$ m) achieves significantly lower FPR ($<$3\%) compared to coarser resolution ($\sim$4--7\%), confirming that boundary approximation errors dominate false positives. Insertion rate decreases logarithmically as the map saturates, with both resolutions converging to similar final rates ($\sim$7\%), indicating consistent orientation coverage density.}
\label{fig:multi_robot_comparison}
\end{figure}

Table~\ref{tab:ikflow_verification} further examines the effect of spatial resolution on our method. Comparing $\Delta=0.05$ m and $\Delta=0.02$ m reveals several key observations:

\paragraph{\textbf{Spatial resolution effect on FPR.}} The most significant difference between $\Delta=0.05$ m and $\Delta=0.02$ m is the dramatic FPR reduction by about $2.5\times$ improvement. 
This observation confirms our hypothesis that boundary approximation errors dominate FPR: finer cells better approximate the irregular reachability boundaries. It also underscores the importance of identifying optimal resolution parameters for balancing prediction accuracy and computational efficiency.

    
\paragraph{\textbf{Map size scaling.}} Transitioning from $\Delta=0.05$ m to $\Delta=0.02$ m increases total inserted poses by $\sim$13--14$\times$ (e.g., Panda: 11.3M $\rightarrow$ 160.5M, UR5e: 20.5M $\rightarrow$ 260.0M). This scaling reflects both the $15.6\times$ increase in grid cells and slightly varying occupancy rates across resolutions. Despite the larger map size, the average poses per cell remains relatively stable (Panda: 503 $\rightarrow$ 490, iiwa7: 367 $\rightarrow$ 358), indicating consistent orientation coverage density.
    
\paragraph{\textbf{Robot-specific workspace characteristics.}} UR5e consistently exhibits the largest workspace coverage (occupancy $\sim$50--53\%), which correlates with its higher FPR, as more boundary voxels accumulate approximation errors. Conversely, Panda and Kinova3 show lower occupancy ($\sim$32--35\%), reflecting more constrained joint configurations.
    
    
\paragraph{\textbf{Accuracy-resolution trade-off.}} While finer resolution ($\Delta=0.02$ m) offers superior FPR and overall accuracy ($\sim$0.984 vs. $\sim$0.978), it requires $\sim$14$\times$ more storage and proportionally longer construction time. Query time also increases to $\sim$13--15 $\mu$s due to larger data/cache overheads, but remains $3\times$ faster than the baseline RM4D. For applications tolerating FPR $<$5\%, $\Delta=0.05$ m provides a cost-effective balance. Critical applications demanding minimal false positives should adopt $\Delta=0.02$ m and insert more.
\subsection{Similarity-Guided Diffusion Policy Transfer: Block Pushing Experiment}
\label{subsec:exp3}

We leverage the robot similarity metric for energy-based guidance in diffusion policy~\cite{chi2023diffusionpolicy,janner2022planning}. The discrete MMD grid in $\mathbb{R}^3$ space with boundaries $\mathcal{W}$ from Section~\ref{subsec:applications} is transformed into a smooth energy landscape through three stages:

\textbf{Stage 1: Seed initialization.} Cells are classified as \textit{seed cells} (high similarity, fixed at zero energy) or \textit{unfilled cells} based on the MMD threshold:
\begin{equation}
E_c^{(0)} = \begin{cases}
0 & \text{if } s_c \leq \tau \quad \text{(seed cell, fixed throughout)} \\
\text{unfilled} & \text{otherwise}
\end{cases}
\end{equation}
where $\tau$ is the similarity threshold (set as the mean of all positive MMD values). Additionally, the central region near the robot base and any isolated zero-energy cells (with no filled neighbors) are smoothed as unfilled to avoid spurious seeds.

\textbf{Stage 2: Wavefront energy propagation.} Energy propagates outward from seed cells layer by layer. At each iteration $t$, the \textit{boundary set} $\mathcal{B}^{(t)}$ consists of all currently unfilled cells that have at least one filled 6-connected neighbor. Each boundary cell is then updated:
\begin{equation}
E_c^{(t+1)} = \frac{1}{|\mathcal{N}_c^{*}|} \sum_{c' \in \mathcal{N}_c^{*}} E_{c'}^{(t)} + \delta, \quad \forall \, c \in \mathcal{B}^{(t)},
\end{equation}
where $\mathcal{N}_c^{*}$ is the subset of connected neighbors of $c$ that already have assigned energy ($E_{c'} \geq 0$), and $\delta = 0.01$ is a fixed increment. Crucially, only cells in $\mathcal{B}^{(t)}$ are updated---seed cells and previously filled cells retain their values unchanged. Propagation \textbf{terminates} when $\mathcal{B}^{(t)} = \emptyset$, i.e., every cell has been assigned an energy value only once.

\textbf{Stage 3: Smoothing.} A 3D Gaussian filter smooths the energy landscape to yield a differentiable field $E: \mathbb{R}^3 \rightarrow \mathbb{R}$, where lower energy indicates higher robot similarity.

Following the energy-based model framework~\cite{yu2020training,xu2024energy}, we interpret energy as unnormalized log-probability:
\begin{equation}
p(\mathbf{o}_\mathbf{x}) \propto e^{-E(\mathbf{o}_\mathbf{x})}, \quad \text{thus} \quad \nabla \log p(\mathbf{o}_\mathbf{x}) = -\nabla E(\mathbf{o}_\mathbf{x}),
\end{equation}
where $\mathbf{o}_\mathbf{x} \in \mathbb{R}^3$ represents the end-effector pose position. For a diffusion model generating robot trajectories with predicted noise $\epsilon(\mathbf{a}|\mathbf{o})$, we incorporate reachability guidance:
\begin{equation}
\bar{\epsilon}(\mathbf{a}|\mathbf{o}) = \epsilon(\mathbf{a}|\mathbf{o}) + \lambda \, \nabla E(\mathbf{o}_\mathbf{x}),
\end{equation}
where $\lambda$ is the guidance strength. This formulation biases generated trajectories toward workspace regions with higher reachability similarity, enabling effective policy transfer.

\subsubsection{Experimental Setup.}
To validate the practical effectiveness of reachability-guided policy transfer, we adopted the multi-modal block pushing experiment~\cite{florence2022implicit,shafiullah2022behavior}. The task involves manipulating two blocks to push them into two designated square target regions, with no requirement for color matching between blocks and targets. This challenging manipulation task tests both trajectory feasibility and task competence across robot morphologies.

Following standard evaluation practices, we train policies on the source robot (xArm6) and select the three best checkpoints from three independent training seeds (seeds 42, 43, 44), yielding nine models total for evaluation.

As for policy transfer, we conduct transfer experiments to three Universal Robots from the UR series with different scales: UR3 (compact), UR5 (medium), and UR10 (large). This selection covers significant variation in workspace sizes and kinematic structures, testing the robustness of similarity-based guidance.

Each trained policy is evaluated under two conditions: (1) \textit{Direct transfer} ($\lambda=0.0$): policy predictions used without modification, and (2) \textit{Guided transfer}: energy-based guidance applied with varying strength $\lambda \in [0.1, 0.2, \cdots, 1.0]$. 

\begin{table}[t]
\centering
\caption{Success rates for policy transfer from xArm6 source to UR series robots over 9 models.}
\label{tab:block_pushing_results}
\small
\begin{tabular}{lccccccc}
\toprule
\multirow{2}{*}{Robot} & \multicolumn{2}{c}{Direct Transfer} & \multicolumn{2}{c}{Guided Transfer} & \multirow{2}{*}{Improvement} \\
\cmidrule(lr){2-3} \cmidrule(lr){4-5}
 & Success & Range & Success & Range & \\
\midrule
xArm6 (source) & 95.40\% & 92.94--98.00\% & -- & -- & -- \\
\midrule
UR3 & 64.64\% & 62.82--71.88\% & 91.34\% & 87.84--95.96\% & +26.70\%  \\
UR5 & 95.15\% & 92.86--98.98\% & 94.60\% & 89.88--98.98\% & $-$0.54\% \\
UR10 & 75.96\% & 74.70--82.74\% & 82.72\% & 77.64--86.78\% & +6.76\% \\
\midrule
\textbf{Average} & \textbf{78.58\%} & -- & \textbf{89.56\%} & -- & \textbf{+10.97\%} \\
\bottomrule
\end{tabular}
\end{table}


\begin{figure}[ht]
\centering
\includegraphics[width=\textwidth]{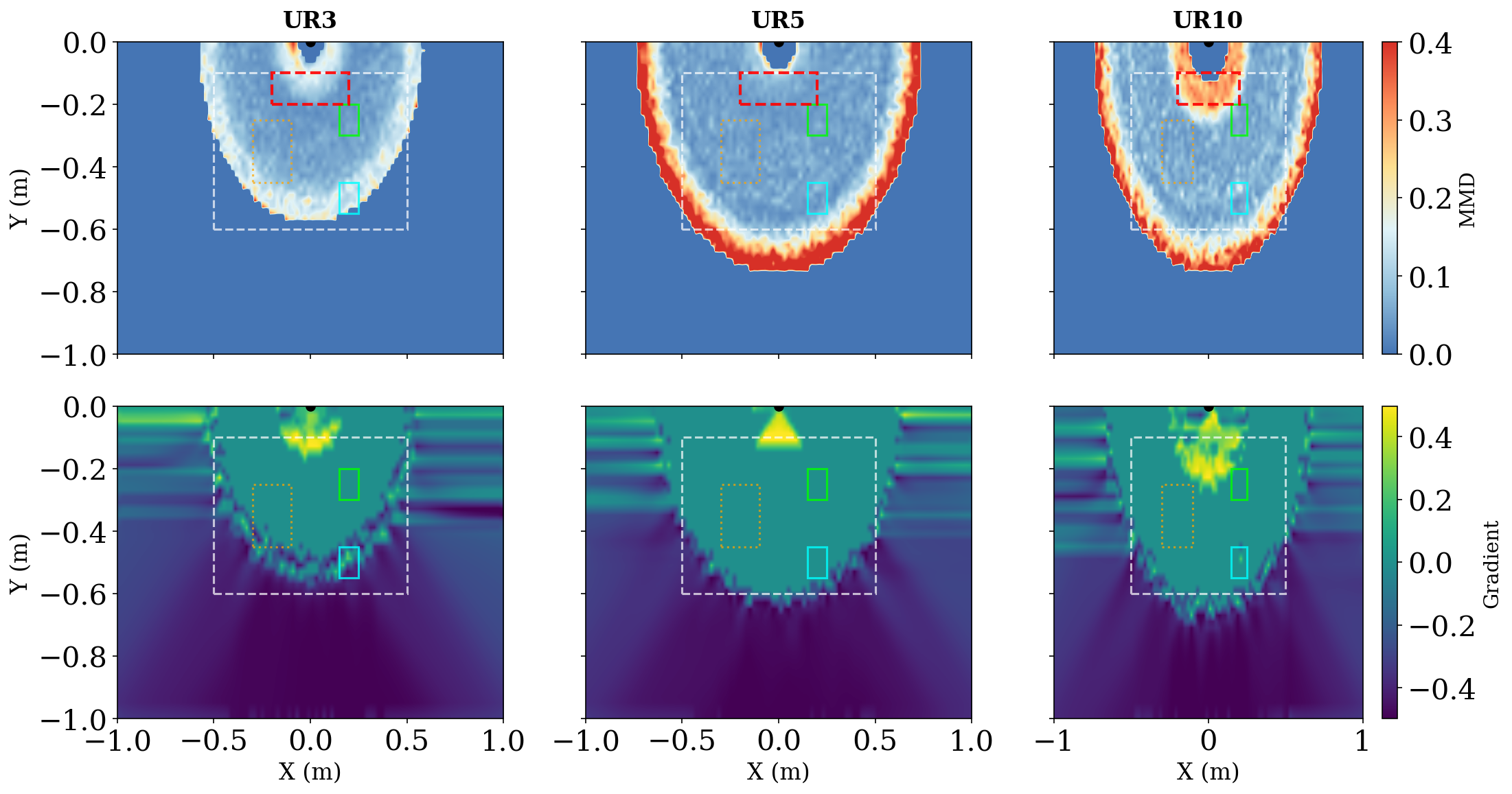}
\caption{Top view field visualizations across UR series transfers. Rows show similarity (top, via MMD) and gradient magnitude (bottom, $\|\nabla E\|$), columns show target robots. Similarity colormap: blue indicates low MMD (high similarity), red indicates high MMD (low similarity). The red dashed rectangle marks the \textit{dangerous region} near the robot base where UR3 and UR10 exhibit low similarity (high MMD, orange/red), prone to generating self-constrained poses during direct transfer. In contrast, UR5 maintains high similarity (low MMD, blue) in this region, implying better direct policy transfer performance. Gradient fields (bottom) show how the direction and magnitude of energy increase, indicating how guidance pushes trajectories away from low-similarity regions. Rectangle annotations denote robot base (dot), table area (white dashed), initial block positions (orange dotted), and target regions (red and green solid).}
\label{fig:transfer_fields}
\end{figure}

\begin{figure}[t]
\centering
\includegraphics[width=\textwidth]{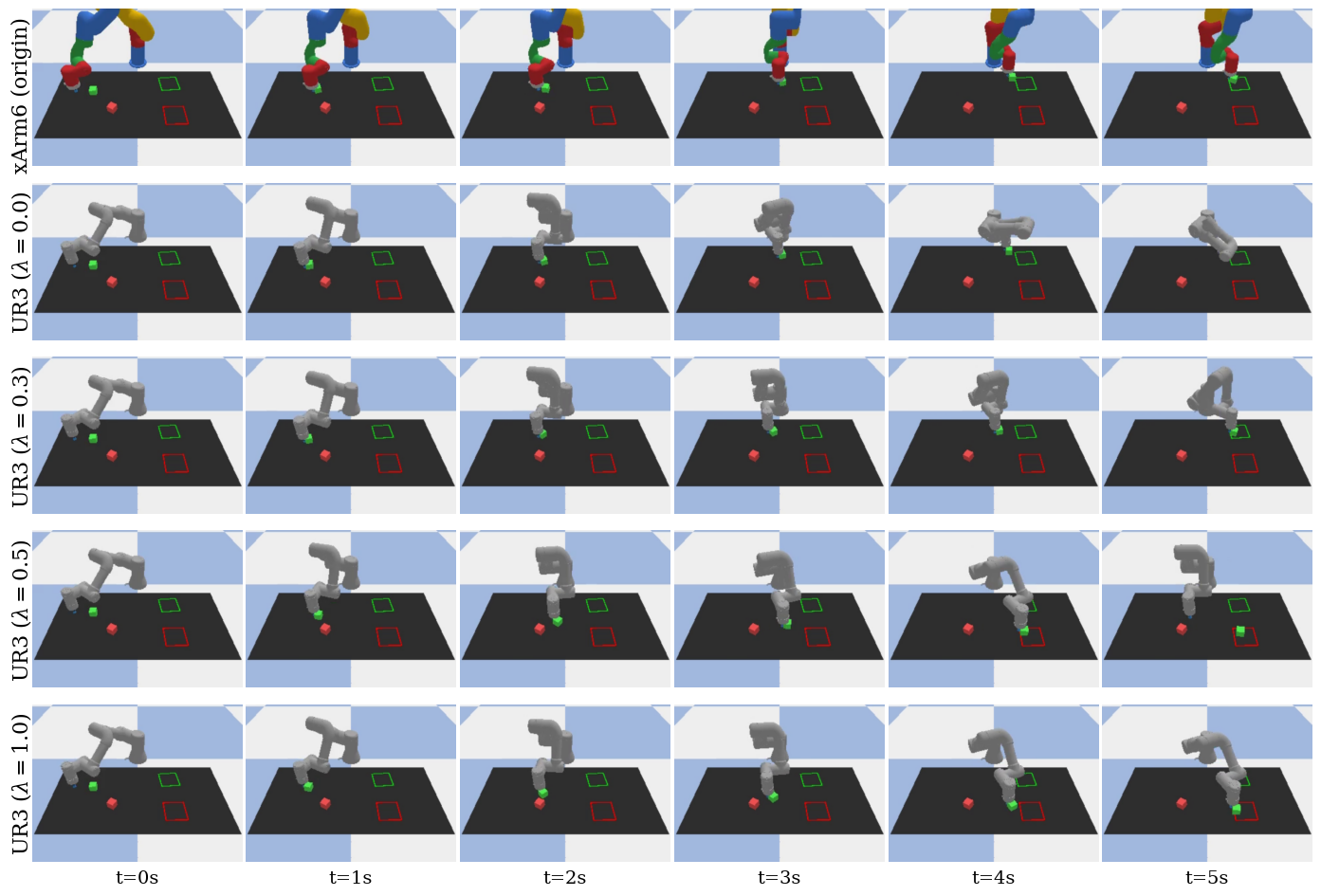}
\caption{Qualitative results showing the block pushing task execution. Top row depicts the xArm6 source policy execution. Subsequent rows show the same task transferred to UR3 with varying guidance strengths ($\lambda = 0.0$ direct transfer, $\lambda = 0.3, 0.5, 1.0$ with increasing guidance). The sequence spans the first 5 seconds at regular time intervals. The block is progressively pushed toward the farther side of the target regions as guidance strength increases, illustrating how reachability guidance influences the trajectory generation.}
\label{fig:video_frames}
\end{figure}

\subsubsection{Results.}
Table \ref{tab:block_pushing_results} summarizes the average success rates across nine trained models for direct transfer (baseline) and guided transfer. The reachability-guided approach achieves an average success rate of 89.56\%, representing a 10.97\% absolute improvement over direct transfer (78.58\%). Notably, the improvement is highly robot-dependent: UR3 and UR10 show substantial gains (+26.70\% and +6.76\%), while UR5 exhibits minimal change ($-$0.54\%).

\paragraph{\textbf{Reachability similarity and transfer performance.}}
The robot-dependent transfer outcomes strongly correlate with workspace reachability similarity patterns revealed in Figure~\ref{fig:transfer_fields}. We identify a critical region near the robot base ($[-0.2, 0.2] \times [-0.2, -0.1]$ m, marked by red dashed rectangle) where kinematic differences between xArm6 and UR robots manifest most prominently. 

UR3 and UR10 exhibit substantial reachability divergence from xArm6 in this region, with large MMD values (visualized as orange and red in the similarity field). When executing xArm6-trained trajectories, these robots frequently generate poses in this region that lead to self-constraints (presented in the second row of Figure~\ref{fig:video_frames}), explaining their reduced baseline success rates of 64.64\% and 75.96\%. The kinematic incompatibility from different robot structure makes these two robots struggle to replicate xArm6's actions near the base.

In contrast, UR5 maintains high reachability similarity throughout this critical region (blue coloring), indicating strong kinematic compatibility with xArm6. This alignment enables UR5 to execute source trajectories directly without modification, achieving 95.15\% baseline success comparable to the source robot's 95.40\%. The near-identical performance validates that reachability similarity serves as a reliable predictor of cross-embodiment transfer feasibility.

\paragraph{\textbf{Energy-based guidance mechanism.}}
The gradient field $\nabla E$ (Figure~\ref{fig:transfer_fields}, bottom row) encodes the direction of increasing energy (decreasing similarity), providing spatial guidance to steer trajectories toward high-similarity regions. During diffusion denoising, this gradient is injected into the predicted noise according to Equation (14), biasing the iterative refinement process.

Figure~\ref{fig:video_frames} illustrates the progressive effect of guidance strength $\lambda$ on UR3 policy execution. Without guidance ($\lambda=0.0$), the robot follows xArm6's trajectory distribution, frequently entering the low-similarity region near the base and failing the task. As $\lambda$ increases from 0.3 to 1.0, the gradient pushes trajectories away from this region with increasing strength. 

The robot adapts by preferentially moving or pushing blocks toward the target region farther from the base, avoiding problematic arm configurations. This behavioral shift demonstrates how energy-based guidance exploits the multimodal property of diffusion policies---rather than forcing the robot to execute infeasible trajectories, it redirects the policy to explore alternative, kinematically feasible modes of task completion.  

The improvements of UR3 and UR10 reflect the effects of bridging the kinematic gap between xArm6 and these robots. Conversely, UR5's minimal change ($-$0.54\%) is expected---since the critical region already exhibits high similarity, guidance provides no additional benefit and may introduce minor perturbations that occasionally degrade performance.

    
    

\section{Conclusion}
This paper introduced RichMap, a grid-based reachability map that simplifies the classical spatial-rotational structure for efficient storage and insertion, while achieving higher precision through careful assumptions and theoretical analysis.
Empirical validation confirms the design: across four robot manipulators, our map achieves comparable accuracy to RM4D (within 0.5\% FPR) while consistently outperforming Capability Maps in false positive rates (25--35\% FPR reduction), demonstrating that we can achieve performance close to compact forms while maintaining the flexibility of the grid-based structure. Moreover, our asynchronous pipeline and GPU acceleration enable large batched query speeds $\sim$18$\times$ faster than sequential counterparts. With finer spatial discretization (0.02 m), our method further reduces FPR to $\sim$1.6\%. The modular structure supports extended applications, including robot similarity analysis and energy-based guidance for diffusion policy transfer. In a multimodal block pushing experiment, the guided policy transfer strategy demonstrates an average 11\% improvement across three target robots, with particularly strong gains for kinematically dissimilar embodiments.

\subsubsection{Limitations and Future Work.}
Our current similarity-based guidance operates only on 3D positions, which may be insufficient for orientation-sensitive tasks.
Future work aims to explore more elegant integration of reachability maps in more challenging and general learning-based manipulation tasks, such as constructing full $SE(3)$ similarity, combining with other values (e.g., collision fields), trajectory planning within safe reachable set~\cite{bansal2017hamilton,holmes2020reachable}, and leveraging motion similarity analysis~\cite{makondo2015knowledge,urain2019generalized,zielinska2023measure} to capture dynamic properties based on static reachability.
Neural network compression could also provide faster prediction and lighter memory usage while preserving the map's utility.

\subsubsection{\ackname}


This work extensively involves GitHub Copilot for both code building and documentation drafting except bibliography. 
Careful line-by-line review, editing, adjustments, and reconstruction with steady buildup strategy for the main structure were adopted for correctness and clarity, supported by rigorous theoretical proofs. 
Detailed testing and cross-validation were performed to verify the reliability and reproducibility of all results related to reachability map construction.

This work was supported by the Natural Science Foundation of China (62461160309), the NSFC-RGC Joint Research Scheme (N\_HKU705/24),  Hong Kong RGC (GRF 17201025, GRF17200924). Yupu Lu is sponsored by the HKU Presidential PhD Scholarship.

%
%
%
\bibliographystyle{splncs04}
\bibliography{richmap}

@article{jiang2025graph,
  title={Graph Reinforcement Learning-based Reachability Map for Generalized Mobile Manipulation},
  author={Jiang, Lu and Ren, Junkai and Zhou, Zhiqian and Qu, Yuke and Lu, Huimin and Wu, Meiping},
  journal={IEEE Transactions on Cognitive and Developmental Systems},
  year={2025},
  publisher={IEEE}
}

@article{cavelli2025modeling,
  title={Modeling the reachability space of robotic manipulators through ellipsoid equations},
  author={Cavelli, Rosario Francesco and Cen Cheng, Pangcheng David and Indri, Marina},
  journal={Journal of Intelligent \& Robotic Systems},
  volume={111},
  number={3},
  pages={90},
  year={2025},
  publisher={Springer}
}

@INPROCEEDINGS{Rudorfer2025RM4D,
  author={Rudorfer, Martin},
  booktitle={2025 IEEE International Conference on Robotics and Automation (ICRA)}, 
  title={RM4D: A Combined Reachability and Inverse Reachability Map for Common 6-/7-Axis Robot Arms by Dimensionality Reduction to 4D}, 
  year={2025},
  volume={},
  number={},
  pages={7689-7695},
  doi={10.1109/ICRA55743.2025.11128095}
}

@article{quan2022dexterity,
  title={The dexterity capability map for a seven-degree-of-freedom manipulator},
  author={Quan, Yuan and Zhao, Chong and Lv, Congmin and Wang, Ke and Zhou, Yanlin},
  journal={Machines},
  volume={10},
  number={11},
  pages={1038},
  year={2022},
  publisher={MDPI}
}

@inproceedings{han2021efficient,
  title={Efficient se (3) reachability map generation via interplanar integration of intra-planar convolutions},
  author={Han, Yiheng and Pan, Jia and Xia, Mengfei and Zeng, Long and Liu, Yong-Jin},
  booktitle={2021 IEEE International Conference on Robotics and Automation (ICRA)},
  pages={1854--1860},
  year={2021},
  organization={IEEE}
}

@article{zacharias2013capability,
  title={The capability map: A tool to analyze robot arm workspaces},
  author={Zacharias, Franziska and Borst, Christoph and Wolf, Sebastian and Hirzinger, Gerd},
  journal={International Journal of Humanoid Robotics},
  volume={10},
  number={04},
  pages={1350031},
  year={2013},
  publisher={World Scientific}
}

@article{cao2011accurate,
  title={Accurate numerical methods for computing 2d and 3d robot workspace},
  author={Cao, Yi and Lu, Ke and Li, Xiujuan and Zang, Yi},
  journal={International Journal of Advanced Robotic Systems},
  volume={8},
  number={6},
  pages={76},
  year={2011},
  publisher={SAGE Publications Sage UK: London, England}
}

@inproceedings{porges2015reachability,
  title={Reachability and dexterity: Analysis and applications for space robotics},
  author={Porges, Oliver and Lampariello, Roberto and Artigas, Jordi and Wedler, Armin and Borst, Christoph and Roa, M{\'a}ximo A},
  booktitle={Workshop on advanced space technologies for robotics and automation-ASTRA},
  year={2015}
}

@article{castelli2008fairly,
  title={A fairly general algorithm to evaluate workspace characteristics of serial and parallel manipulators},
  author={Castelli, Gianni and Ottaviano, Erika and Ceccarelli, Marco},
  journal={Mechanics based design of structures and machines},
  volume={36},
  number={1},
  pages={14--33},
  year={2008},
  publisher={Taylor \& Francis}
}

@inproceedings{zacharias2007capturing,
  title={Capturing robot workspace structure: representing robot capabilities},
  author={Zacharias, Franziska and Borst, Christoph and Hirzinger, Gerd},
  booktitle={2007 IEEE/RSJ International Conference on Intelligent Robots and Systems},
  pages={3229--3236},
  year={2007},
  organization={Ieee}
}

@article{yao2023enhanced,
  title={Enhanced dexterity maps (edm): A new map for manipulator capability analysis},
  author={Yao, Haowen and Laha, Riddhiman and Figueredo, Luis FC and Haddadin, Sami},
  journal={IEEE Robotics and Automation Letters},
  volume={9},
  number={2},
  pages={1628--1635},
  year={2023},
  publisher={IEEE}
}

@article{yoshikawa1985manipulability,
  title={Manipulability of robotic mechanisms},
  author={Yoshikawa, Tsuneo},
  journal={The international journal of Robotics Research},
  volume={4},
  number={2},
  pages={3--9},
  year={1985},
  publisher={Sage Publications Sage CA: Thousand Oaks, CA}
}

@article{jauhri2022robot,
  title={Robot learning of mobile manipulation with reachability behavior priors},
  author={Jauhri, Snehal and Peters, Jan and Chalvatzaki, Georgia},
  journal={IEEE Robotics and Automation Letters},
  volume={7},
  number={3},
  pages={8399--8406},
  year={2022},
  publisher={IEEE}
}

@article{jiang2022learning,
  title={Learning suction graspability considering grasp quality and robot reachability for bin-picking},
  author={Jiang, Ping and Oaki, Junji and Ishihara, Yoshiyuki and Ooga, Junichiro and Han, Haifeng and Sugahara, Atsushi and Tokura, Seiji and Eto, Haruna and Komoda, Kazuma and Ogawa, Akihito},
  journal={Frontiers in Neurorobotics},
  volume={16},
  pages={806898},
  year={2022},
  publisher={Frontiers Media SA}
}

@inproceedings{kluge2021mobile,
  title={Mobile robot base placement for assembly systems: survey, measures and task clustering},
  author={Kluge-Wilkes, Aline and Schmitt, Robert H},
  booktitle={Congress of the German Academic Association for Production Technology},
  pages={439--447},
  year={2021},
  organization={Springer}
}

@article{zhang2020novel,
  title={A novel coordinated motion planner based on capability map for autonomous mobile manipulator},
  author={Zhang, Heng and Sheng, Qi and Sun, Yuxin and Sheng, Xinjun and Xiong, Zhenhua and Zhu, Xiangyang},
  journal={Robotics and autonomous systems},
  volume={129},
  pages={103554},
  year={2020},
  publisher={Elsevier}
}

@inproceedings{makhal2018reuleaux,
  title={Reuleaux: robot base placement by reachability analysis},
  author={Makhal, Abhijit and Goins, Alex K},
  booktitle={2018 second IEEE international conference on robotic computing (IRC)},
  pages={137--142},
  year={2018},
  organization={IEEE}
}

@inproceedings{burget2015stance,
  title={Stance selection for humanoid grasping tasks by inverse reachability maps},
  author={Burget, Felix and Bennewitz, Maren},
  booktitle={2015 IEEE International conference on robotics and automation (ICRA)},
  pages={5669--5674},
  year={2015},
  organization={IEEE}
}

@inproceedings{vahrenkamp2013robot,
  title={Robot placement based on reachability inversion},
  author={Vahrenkamp, Nikolaus and Asfour, Tamim and Dillmann, R{\"u}diger},
  booktitle={2013 IEEE International Conference on Robotics and Automation},
  pages={1970--1975},
  year={2013},
  organization={IEEE}
}

@inproceedings{bansal2017hamilton,
  title={Hamilton-jacobi reachability: A brief overview and recent advances},
  author={Bansal, Somil and Chen, Mo and Herbert, Sylvia and Tomlin, Claire J},
  booktitle={2017 IEEE 56th annual conference on decision and control (CDC)},
  pages={2242--2253},
  year={2017},
  organization={IEEE}
}

@inproceedings{holmes2020reachable,
  title={Reachable sets for safe, real-time manipulator trajectory design},
  author={Holmes, Patrick and Kousik, Shreyas and Zhang, Bohao and Raz, Daphna and Barbalata, Corina and Johnson-Roberson, Matthew and Vasudevan, Ram},
  booktitle={Proceedings of Robotics: Science and Systems (RSS)},
  year={2020}
}

@article{xu2021optimal,
  title={Optimal grasping pose for dual-arm space robot cooperative manipulation based on global manipulability},
  author={Xu, Ruonan and Luo, Jianjun and Wang, Mingming},
  journal={Acta Astronautica},
  volume={183},
  pages={300--309},
  year={2021},
  publisher={Elsevier}
}

@article{vahrenkamp2015representing,
  title={Representing the robot’s workspace through constrained manipulability analysis},
  author={Vahrenkamp, Nikolaus and Asfour, Tamim},
  journal={Autonomous Robots},
  volume={38},
  number={1},
  pages={17--30},
  year={2015},
  publisher={Springer}
}

@inproceedings{vahrenkamp2012manipulability,
  title={Manipulability analysis},
  author={Vahrenkamp, Nikolaus and Asfour, Tamim and Metta, Giorgio and Sandini, Giulio and Dillmann, R{\"u}diger},
  booktitle={2012 12th ieee-ras international conference on humanoid robots (humanoids 2012)},
  pages={568--573},
  year={2012},
  organization={IEEE}
}

@article{ames2022ikflow,
  title={Ikflow: Generating diverse inverse kinematics solutions},
  author={Ames, Barrett and Morgan, Jeremy and Konidaris, George},
  journal={IEEE Robotics and Automation Letters},
  volume={7},
  number={3},
  pages={7177--7184},
  year={2022},
  publisher={IEEE}
}

@article{sandakalum2022motion,
  title={Motion planning for mobile manipulators—a systematic review},
  author={Sandakalum, Thushara and Ang Jr, Marcelo H},
  journal={Machines},
  volume={10},
  number={2},
  pages={97},
  year={2022},
  publisher={MDPI}
}

@inproceedings{makondo2015knowledge,
  title={Knowledge transfer for learning robot models via local procrustes analysis},
  author={Makondo, Ndivhuwo and Rosman, Benjamin and Hasegawa, Osamu},
  booktitle={2015 IEEE-RAS 15th International Conference on Humanoid Robots (Humanoids)},
  pages={1075--1082},
  year={2015},
  organization={IEEE}
}

@inproceedings{urain2019generalized,
  title={Generalized multiple correlation coefficient as a similarity measurement between trajectories},
  author={Urain, Julen and Peters, Jan},
  booktitle={2019 IEEE/RSJ International Conference on Intelligent Robots and Systems (IROS)},
  pages={1363--1369},
  year={2019},
  organization={IEEE}
}

@article{zielinska2023measure,
  title={The measure of motion similarity for robotics application},
  author={Zielinska, Teresa and Coba, Gabriel},
  journal={Sensors},
  volume={23},
  number={3},
  pages={1643},
  year={2023},
  publisher={MDPI}
}

@article{gretton2006kernel,
  title={A kernel method for the two-sample-problem},
  author={Gretton, Arthur and Borgwardt, Karsten and Rasch, Malte and Sch{\"o}lkopf, Bernhard and Smola, Alex},
  journal={Advances in neural information processing systems},
  volume={19},
  year={2006}
}

@article{gretton2012kernel,
  title={A kernel two-sample test},
  author={Gretton, Arthur and Borgwardt, Karsten M and Rasch, Malte J and Sch{\"o}lkopf, Bernhard and Smola, Alexander},
  journal={The journal of machine learning research},
  volume={13},
  number={1},
  pages={723--773},
  year={2012},
  publisher={JMLR. org}
}

@inproceedings{florence2022implicit,
  title={Implicit behavioral cloning},
  author={Florence, Pete and Lynch, Corey and Zeng, Andy and Ramirez, Oscar A and Wahid, Ayzaan and Downs, Laura and Wong, Adrian and Lee, Johnny and Mordatch, Igor and Tompson, Jonathan},
  booktitle={Conference on robot learning},
  pages={158--168},
  year={2022},
  organization={PMLR}
}

@article{shafiullah2022behavior,
  title={Behavior transformers: Cloning $ k $ modes with one stone},
  author={Shafiullah, Nur Muhammad and Cui, Zichen and Altanzaya, Ariuntuya Arty and Pinto, Lerrel},
  journal={Advances in neural information processing systems},
  volume={35},
  pages={22955--22968},
  year={2022}
}

@inproceedings{janner2022planning,
  title={Planning with Diffusion for Flexible Behavior Synthesis},
  author={Janner, Michael and Du, Yilun and Tenenbaum, Joshua and Levine, Sergey},
  booktitle={International Conference on Machine Learning},
  pages={9902--9915},
  year={2022},
  organization={PMLR}
}

@inproceedings{chi2023diffusionpolicy,
  title={Diffusion Policy: Visuomotor Policy Learning via Action Diffusion},
  author={Chi, Cheng and Feng, Siyuan and Du, Yilun and Xu, Zhenjia and Cousineau, Eric and Burchfiel, Benjamin and Song, Shuran},
  booktitle={Proceedings of Robotics: Science and Systems (RSS)},
  year={2023}
}

@inproceedings{yu2020training,
  title={Training deep energy-based models with f-divergence minimization},
  author={Yu, Lantao and Song, Yang and Song, Jiaming and Ermon, Stefano},
  booktitle={International Conference on Machine Learning},
  pages={10957--10967},
  year={2020},
  organization={PMLR}
}

@article{xu2024energy,
  title={Energy-based diffusion language models for text generation},
  author={Xu, Minkai and Geffner, Tomas and Kreis, Karsten and Nie, Weili and Xu, Yilun and Leskovec, Jure and Ermon, Stefano and Vahdat, Arash},
  journal={arXiv preprint arXiv:2410.21357},
  year={2024}
}
\end{document}